\documentclass{article}

\usepackage{microtype}
\usepackage{graphicx}
\usepackage{subfigure}
\usepackage{booktabs} 

\usepackage{hyperref}

\usepackage[accepted]{icml2024}

\usepackage{amsmath}
\usepackage{amssymb}
\usepackage{mathtools}
\usepackage{amsthm}

\usepackage[capitalize,noabbrev]{cleveref}

\theoremstyle{plain}

\theoremstyle{definition}

\theoremstyle{remark}

\usepackage[textsize=tiny]{todonotes}

\icmltitlerunning{Are Synthetic Time-series Data Really not as Good as Real Data?}

\begin{document}

\twocolumn[

\icmltitle{Are Synthetic Time-series Data Really not as Good as Real Data?}







\begin{icmlauthorlist}
\icmlauthor{Fanzhe Fu}{zju}
\icmlauthor{Junru Chen}{zju}
\icmlauthor{Jing Zhang}{ruc}
\icmlauthor{Carl Yang}{emory}
\icmlauthor{Lvbin Ma}{comp}
\icmlauthor{Yang Yang}{zju}
\end{icmlauthorlist}

\icmlaffiliation{zju}{College of Computer Science and Technology, Zhejiang University, Hangzhou, China}
\icmlaffiliation{ruc}{School of Information, Renmin University of China, Haidian District Beijing, China}
\icmlaffiliation{emory}{Department of Computer Science, Emory University, Atlanta, United States}
\icmlaffiliation{comp}{Zhejiang Huayun Information Technology Ltd., Hangzhou, China}


\icmlcorrespondingauthor{Yang Yang}{yangya@zju.edu.cn}

\icmlkeywords{Machine Learning, ICML}

\vskip 0.3in
]



\printAffiliationsAndNotice{}  


\begin{abstract}
Time-series data presents limitations stemming from data quality issues, bias and vulnerabilities, and generalization problem. Integrating universal data synthesis methods holds promise in improving generalization. However, current methods cannot guarantee that the generator's output covers all unseen real data. In this paper, we introduce InfoBoost-- a highly versatile cross-domain data synthesizing framework with time series representation learning capability. We have developed a method based on synthetic data that enables model training without the need for real data, surpassing the performance of models trained with real data. Additionally, we have trained a universal feature extractor based on our synthetic data that is applicable to all time-series data. Our approach overcomes interference from multiple sources rhythmic signal, noise interference, and long-period features that exceed sampling window capabilities. Through experiments, our non-deep-learning synthetic data enables models to achieve superior reconstruction performance and universal explicit representation extraction without the need for real data. 

\end{abstract}

\begin{figure*}[!htbp] 
  \centering
  \includegraphics[width=1.0\linewidth]{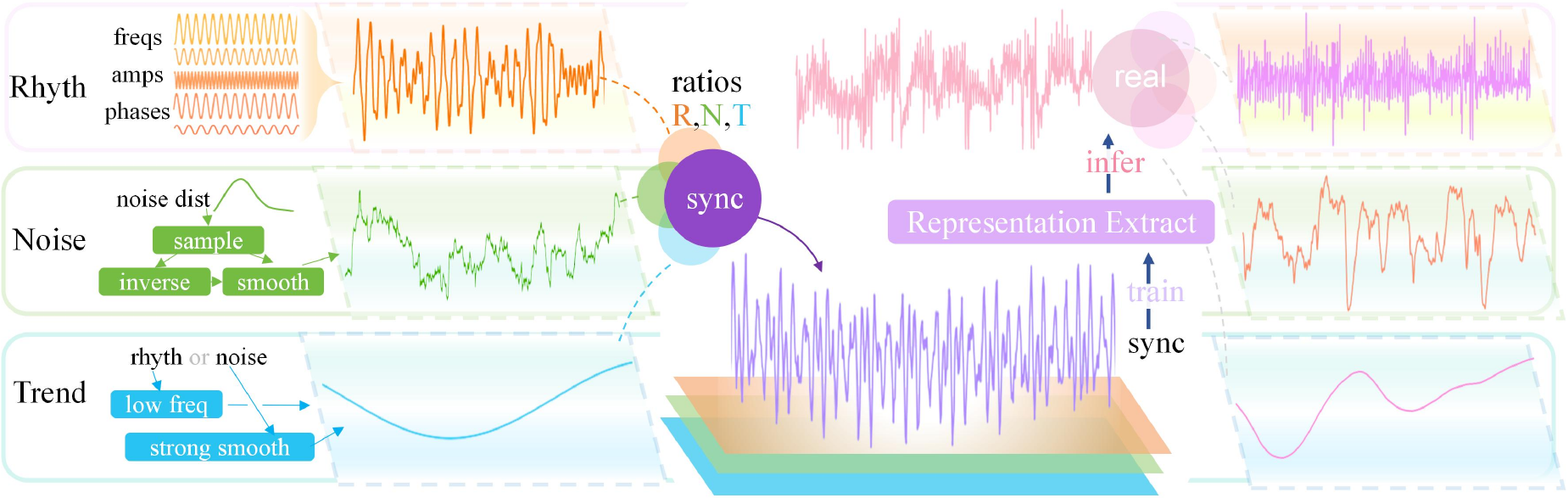}
  \vspace{-5mm}
  \caption{Schematic of the InfoBoost model illustrating the data synthesis, representation learning, and representation prediction processes. In the diagram, the label 'Rhyth' corresponds to the multi-source rhythmic data (MRD), 'Noise' corresponds to different types of noise and their noise ratios (TN \& NR), and 'Trend' corresponds to trend information (TI). This visual representation elucidates the individual roles of each component within the InfoBoost framework. 'Sync' stands for 'synthesized data', representing data that is artificially generated, integrating MRD, TN \& NR, and TI.
}
  \label{InfoBoost_modelGraph}
  \vspace{-2mm}
\end{figure*}

\section{Introduction}
\label{Introduction}
There are widespread applications of time-series data, including financial analysis \cite{TANG2022363_financeTs}, energy data analysis \cite{Sriramalakshmi2022_powerTs},  traffic flow prediction \cite{SHAYGAN2022103921_trafficTs}, weather data analysis \cite{sym15040951_weatherTs}, and physiological signal analysis in healthcare domain\cite{saeidi2021neural_eegTs, wang2022systematic_medicalTs, zhang2023brant_iEEGTS}. Unlike large language models (LLM) which can integrate and train models with diverse domain data so as to significantly enhance their generalization and adaptability \cite{zhao2023survey_LLM}, when applying general time-series large models to various types of tasks involving time-series data, limitations inherent to time-series data inevitably appear: 
(1) Data Quality Issues: The fine-tuning process for temporal data needs to be handled carefully as it may contain adversarial or noisy examples, which could impact the model’s robustness; (2) Bias and Vulnerabilities: The use of temporal data may cause the model to inherit biases or vulnerabilities from the data, thereby reducing its robustness in real-world applications; (3) Generalization Problems: Despite being trained on vast datasets, time-series models may not generalize well to unseen or out-of-distribution data. Time-series and spatio-temporal data may exhibit sudden shifts or trends, potentially leading to unreliable outputs, highlighting the need for robust generalization \cite{jin2023largeTsLimit}.

To address the limitations of real-world time-series data, inspired by the potential benefits of synthetic data \cite{Savage2023_whySync, luo2023time_augments, Yin2023_whySync, zhang2024resimad_whySync}, we explored synthetic time-series data generation methods. However, we observed that almost all time-series data generation approaches require sampling training data from real datasets or fine-tuning deep learning-based generators \cite{luo2023time_augments, knosys_2022_relate, yang2022unsupervised_relate}. While existing methods can somewhat alleviate data quality and bias and vulnerabilities issues, they still cannot guarantee that the generator's output encompasses unseen data, thereby constraining the generalization of synthetic data. Therefore, we believe that to address the limitations of synthetic time-series data, specially in terms of generalization, a universal time-series data synthesis method that does not rely on real data is essential.

Our first challenge is to deal with the significant variations presented in time-series data across different datasets or domains. These variations encompass differences in data distribution, time scales, signal-to-noise ratio, and feature types, among others. In order to establish a universally applicable approach for diverse time-series data, we draw inspiration from a classic method for general time series feature extraction: the Fourier transform. Usually, in the field of time series analysis, widely used transformations such as the Continuous Fourier Transform (CFT), Discrete Fourier Transform (DFT), and Discrete Cosine Transform (DCT), all come with equations suggesting that each sample point contains information about the frequency \cite{Zieliński2021_ft}. Therefore, CFT, DFT and DCT transformations convert these sample points into corresponding frequency components in the frequency domain, based on their underlying principle that each sample point in the time-series data can be represented by a set of frequency amplitude and phase. Building upon the fundamental capabilities of the frequency domain transformations, we have designed a method to synthesize data by superimposing several sine waves with varying phases, frequencies and amplitudes, simulating a range of rhythmic signals that may occur in the real world.

However, other challenges arise from the complex nature of time-series data itself. Most real-world time-series data involve complex signal sources or a mixture of multiple sources, and, often containing noise of different frequencies and distributions. Additionally, they may encompass long-term features that cannot be identified in the sampled data because their minimum feature period exceeds the sampling window. Consequently, three more challenges emerge:
(1) Classic frequency domain extraction methods struggle with many real-world time-series data, which often involve multiple signal sources, possibly a mixture of multiple sources, making it difficult to distinguish and independently extract the frequency components of each source.
(2) Time-series data typically contain noise of varying frequencies and distributions, which can interfere with the accurate extraction of frequency components by these transformations. Additionally, these transformations assume that each sample point can be represented by a set of frequency amplitude and phase, which is not always true in real data, especially when the signal is highly complex or the noise level is significant.
(3) Some time-series data may contain long-period features that exceed the range the sampling window can cover, and these transformations may not identify these long-period features in the sampled data, leading to information loss or distortion in the frequency domain.
In conclusion, although these transformations are useful in certain scenarios, their limitations are difficult to avoid when dealing with complex real-world time-series data.

To address these challenges, we develop InfoBoost: a cross-domain highly versatile data synthetic framework with time series representation learning capability. InfoBoost facilitates non-deep-learning data synthesis methods, enabling deep learning models trained solely on synthetic data to outperform those trained on real data. Furthermore, training on synthetic data allows for the creation of a representation extractor applicable across various time series domains without requiring fine-tuning post-synthesis training. The extracted representations from InfoBoost can serve as explicit interpretable features for further visualization analysis and higher-quality frequency domain information extraction.

Our approach to handling the complex composition of time-series data involves the explicit design of separately contained multi-source rhythmic data (MRD) information, various types of noise and their respective noise ratios (TN \& NR), as well as trend information (TI) that extends beyond the sampling window. Real-world data typically lacks these explicit information. To address this, we develop a data synthesis approach that revolves around synthesizing MRD, TN \& NR, and TI to create synthetic data with explicit information. Each set of synthetic data inherently corresponds to a specific set of generating parameters, including MRD, TN \& NR, and TI. The data synthesis process only require sampling methods customized for MRD, TN \& NR, and TI, with various random synthetic parameters. It does not rely on any learnable parameters to generate highly versatile synthetic data to solve the generalization problem of time-series data. This versatility is demonstrated by the fact that a deep learning model trained solely on InfoBoost’s non-deep learning synthetic data outperforms that trained on large amounts of real data when validated on real data test sets, as demonstrated in Section \ref{exp_recon}.

Due to the lacking of MRD, TN \& NR, and TI in real-world data, we utilize synthetic input to train a representation extractor that can extract MRD, TN \& NR, and TI information from the input time-series data. This process, along with the subsequent training phase using synthetic data, can be considered akin to pre-training that relies solely on synthetic data. Due to its ability to capture information across three dimensions—-MRD, TN \& NR, and TI, the trained model can obtain corresponding representations from real data input. InfoBoost’s data synthesis, representation learning, and prediction process illustrated in \autoref{InfoBoost_modelGraph}. The extracted representations from real data can provide explicit analytical and decomposed features for the dataset, offering an interpretable understanding of the time-series data.

Summarized below are the main contributions of this work:

1. We introduce a highly generalizable synthetic data method, emphasizing its capacity to generalize to various real-world datasets solely based on synthetic data, eliminating the need for real data to solve generalization problem.

2. We develop a representation learning method that only relies on synthetic data, enabling the extraction of rhythmic components, noise, and trend information explicitly.

3. We validate the universality of InfoBoost's synthetic data and validate the general representation by visualizing extractor's output across dozens of publicly available datasets, spanning various domains such as finance, weather and healthcare.

\section{Related works}
Although there are a variety of deep learning-based time-series data augmentation and synthesis methods \cite{luo2023time_augments, knosys_2022_relate, yang2022unsupervised_relate}, it is almost impossible to find a method that can be universally applied across domains and does not rely on training deep learning with real data, while simultaneously contributing synthetic data for other machine learning tasks \cite{trirat2024_tsSurvey}. Expanding the probability distribution units for generated data using deep learning-based methods makes it challenging to ensure that the generated data covers unseen or diverse data distribution from other domains, even Meta-learning methods makes assumptions about tasks coming from the same distribution\cite{Swan2022_DLlimit}. The development and exploration of non-deep learning-based universal time-series data synthesis methods provide a promising way to address the limitations of current deep learning-based approaches and improve the generalizability of synthetic data across diverse domains. Notably, our method enables models trained without real data to outperform those trained on real data across all tested datasets, as demonstrated in \autoref{exp_recon}, even in the absence of real data or real data information.

While a wide array of architectural designs \cite{yang2022unsupervised_relate}, loss functions \cite{fraikin2023trep_relate, zhong2023multiscale_relate}, and training methodologies \cite{zerveas2021transformerbased_relate}  have been explored for deep learning-based time series representation learning, current methods neither explicitly separate the MRD, TN \& NR, and TI information across all domains of time-series data as depicted in \autoref{InfoBoost_modelGraph}, nor offer a means of visualizing these information. Unlike existing methods, our work introduces a novel approach for synthesizing explicit features for univariate time-series data without the need to train on real data. Furthermore, we present a general method for decomposing and extracting MRD, TN \& NR, and TI information from time-series data without fine-tuning on real data. The achievement attributes to our data synthesis approach, which constructs MRD, TN \& NR, and TI information based on the parametrization of actual values after random sampling. This approach enables the exact separation of MRD, TN \& NR, and TI information from time-series data, which can serve as labels for training deep learning models.

\section{Methodology}
\subsection{Universal Synthetic time-series data}
\label{sync_data}
In this section, we will demonstrate how to generate multi-source rhythmic data (MRD), different types of noise and their noise ratios (TN \& NR), and trend information (TI) based on parametric design. These components will be combined according to their respective ratios to create synthetic data.

\subsubsection{Generating Multi-source Rhythmic Data}
\label{rhythmic_data}

To ensure that all frequency components of the sine waves have an equal chance of being sampled during the synthesis of rhythms, we calculated the maximum upper and minimum lower limits of the frequencies within the range of random frequency sampling, adhering to the Nyquist-Shannon Sampling Theorem \cite{TAN201913_shannon}. This ensures that all sine waves’ frequencies are uniformly sampled in the synthesis of rhythms. By incorporating classic frequency domain settings, our design involves the creation of rhythmic data comprising both simple and complex waveforms, which constructed from a random number of sin waves with random amplitudes, frequencies, and phases.

The up limit frequency (\(f_{\text{max}}\)) for our rhythm synthesis data is determined in accordance with the principles of the Shannon-Nyquist Sampling Theorem. This theorem ensures that the sampling frequency is at least twice the highest frequency component in the signal to prevent aliasing and to accurately reconstruct the original signal from the sampled data. Additionally, the frequency resolution is influenced by the sampling window length, which in turn affects the main lobe width of the window function and consequently the frequency resolution of the spectral analysis. Taking these principles into consideration, we have established the frequency values for our rhythm synthesis data to ensure appropriate sampling and frequency resolution. Furthermore, the inclusion of \(f_{\text{min}}\) ensures that our frequency parameter aligns with the Nyquist criterion, preventing undersampling and guaranteeing that the synthesized rhythms capture essential low-frequency components, thus maintaining fidelity and accuracy in our synthesized data. The frequency lies within the range shown in Equation \ref{freq_range}, where t typically represents 1 in the context of discrete data points, and N generally denotes the data length, which also known as the sampling window length.

\vspace{-2mm}
\begin{equation}
\label{freq_range}
\begin{aligned}
f_{\text{max}} &= \frac{f_s}{2} = \frac{1}{2t} \\
f_{\text{min}} &\approx \frac{1}{N}
\end{aligned}
\vspace{-2mm}
\end{equation}

The phase values typically range between 0 and \(2\pi\) , aligning with the conventional understanding of the sine function’s periodicity. 

\begin{figure}[htbp] 
  \centering  
  \includegraphics[width=1.0\linewidth]{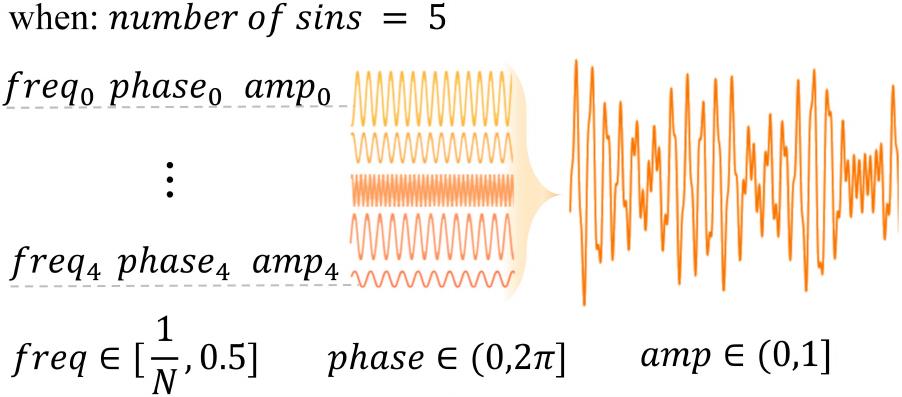}   
  \vspace{-5mm}
  \caption{This image illustrates a possible set of five corresponding sine waves, each obtained by random sampling of frequency phases and amplitudes within their respective ranges, and it should be noted that the number of sine waves is also randomly determined. Additionally, the image showcases the composite rhythmic data generated by the superposition of these randomly determined sine waves.}  
  \label{rhyth_signal}   
  \vspace{-3mm}
\end{figure}

As for amplitudes, they are generally defaulted to fall within the range of 0 to 1. After synthesizing rhythmic data, we randomly sample different numbers of sine waves. Subsequently, upon completion of the sampling process, we normalize the resulting superimposed rhythmic data to the range of 0 to 1. Then the normalization of the amplitudes is computed based on the combined number of sine waves. The synthesis of multi-source rhythmic data is presented in \autoref{rhyth_signal}.

\begin{figure}  
  \centering  
  \vspace{-1mm}
  \includegraphics[width=0.9\linewidth]{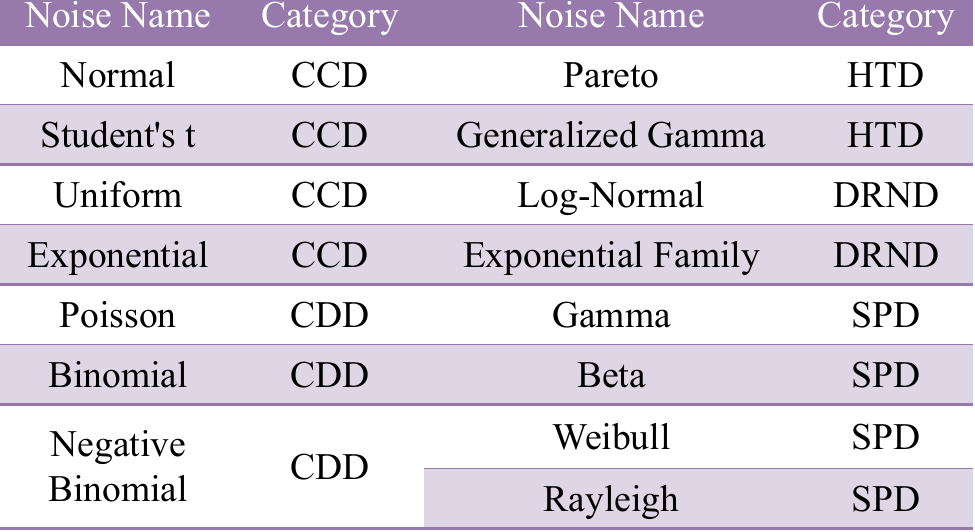}   
  \vspace{-3mm}
  \caption{The list of noise distributions along with their corresponding categories. The "Category" column specifies the category each noise type belongs to. The categories are abbreviated as follows: CCD (Common Continuous Distributions), CDD (Common Discrete Distributions), HTD (Heavy-Tailed Distributions), DRND (Distributions Related to Normal Distribution), and SPD (Shape Parameter Distributions).}  
  \label{noise-table}   
  \vspace{-8.8mm}
\end{figure}  

\subsubsection{Generating Different Types of Noise}
\label{sync_noise}

In order to comprehensively simulate noise information for all types of time-series data, we designed a synthetic noise generator that encompasses 15 different types of noise distributions, classified into 5 primary categories. This design offers a wide range of noise distributions, and meets the demands in simulated outcomes that closely mirror real-world scenarios. To achieve this, for discrete distributions, we adopted Bernoulli \cite{SINHARAY2010132_Discrete}, geometric \cite{SINHARAY2010132_Discrete}, and Poisson distributions \cite{SINHARAY2010132_Discrete}. Additionally, included heavy-tailed distributions \cite{Laplace}, distributions related to the normal distribution (t-distribution \cite{t_distribution} and Pareto distribution \cite{Pareto}), shape parameter distributions (Beta and Gamma distributions \cite{Beta_gamma}), scale parameter distributions (exponential family distribution \cite{Exponential} ) and normal distribution family \cite{Wiley2020_normal}. 15 noise distributions and their 5 categories are listed in  Figure\ref{noise-table}.

Due to the highly uncertain nature of noise distributions in real data, when sampling the noise, we undertake the following three steps: 

1. We customize random parameter sampling based on the fundamental parameters of each noise distribution to introduce relative randomness into each distribution.

2. We perform a partial y-axis inversion on the sampled results of each parameter distribution. This ensures that distributions overly concentrated around a maximum or minimum value do not adversely impact the uniformly normalized sampled results.

3. We apply random kernel size smoothing to the sampled noise, enhancing the diversity of the noise data distribution and simulating temporal dynamics that are difficult to classify as rhythmic information in some real datasets. To prevent overlap between noise and trend information, the kernel size for the random smoothing in the third step is constrained to a relatively small proportion compared to that of the total data length.

\subsubsection{Generating Trend Information}

Our approach to generating trend information involves a randomized selection between two distinct methods. This random selection process is designed to enhance the diversity of the generated data, providing a degree of stochasticity in simulating different types of data trends. By randomly choosing between these two methods, we aim to better emulate the diversity presents in real-world data, where various types of trends and noise patterns are commonly observed. Our trend generation process involves the random selection between two distinct methods:

1. Multi-Sine Trend Generation:
   This method enables the simulation of complex periodic patterns by generating multiple sine waves with random parameters and combining them to form a composite trend. With this approach, we can represent long-period features that exceed the sampling window in real-world phenomena. we utilize a similar approach to the one used in \autoref{rhythmic_data}, which entails the superposition of multiple sine waves. However, unlike \autoref{rhythmic_data}, when generating long-period trends, we constrain the superposition of sine waves to ensure a relatively smaller number of spikes and less complex waveforms in the resulting trend. This adjustment is implemented to better capture the characteristics of long-period trends, ensuring that the generated data aligns with the intended representation of such trends. We ensure that the generated minimum period is greater than the range captured by the sample window. Additionally, we introduce a random multiplier greater than 1 to enhance the diversity of the generated data. This adjustment ensures that the generated periods exhibit a certain level of diversity and can simulate a variety of periodic trends across a wider range.

2. Random Noise Trend Generation:
   This method allows for the simulation of random fluctuations or irregularities often observed in real data, and introduces controlled randomness and smooths out the generated trend, replicating the stochastic nature of many real-world trends. By generating noise with the same method as described in \autoref{sync_noise}, and applying a larger kernel size for smoothing, this approach enhances the diversity and stability of the data, mitigates overfitting, and better simulates real-world data trends.

The random selection between these two methods aims to enhance the diversity and stochasticity of the generated trend data, providing a more realistic representation of the multifaceted nature of data trends commonly observed in real-world datasets.

\subsubsection{Signal-to-Noise-and-Trend Ratio}

Given the rhythm noise and trend information generated through their respective random parameters, the next step involves standardizing each of the three generated outputs to fall within the range of -1 and 1. This standardization facilitates the computation of the contribution ratio of each synthetic component in the final composite data. As this ratio directly determines the signal-to-noise ratio of the rhythmic information in the composite data, it significantly influences the overall performance and characteristics of the synthesized data. We will randomly generate a set of three ratios, whose sum is 1, to serve as the ratios for the rhythmic, noise, and trend components. 
\vspace{-2mm}
\begin{equation}
\label{ratios_sum}
\begin{aligned}
r_{\text{rhyth}} + r_{\text{noise}} + r_{\text{trend}} = 1
\end{aligned}
\vspace{-2mm}
\end{equation}

Consequently, we will utilize these individual ratios to weight the generation of the final composite data, combining the components based on their respective ratios.
\vspace{-2mm}
\begin{equation}
\label{sumToSync}
\begin{aligned}
\text{Sync} = r_{\text{rhyth}} \times \text{Rhyth} + r_{\text{noise}} \times \text{Noise} + r_{\text{trend}} \times \text{Trend}
\end{aligned}
\vspace{-2mm}
\end{equation}

\subsection{Universal Time Series Representation Extraction}
In this section, we will demonstrate how to train a representation extractor solely based on the synthetic data generated from the random parameters and the composite data obtained from \autoref{sync_data}. This representation extractor is designed to explicitly separate MRD, TN \& NR, and TI signals, using only the synthetic data for training.

\subsubsection{Synthetic info normalization}
In data synthesis process, the majority of parameters consist of continuous values, such as frequencies, phases, amplitudes, and ratios. However, there are also crucial parameters that are composed of discrete information, including the number of sine waves, noise type and class, trend class, and similar kernel size parameters resembling smoothing windows, whose actual values may scale significantly with the length of the data sampling window.

\begin{figure}  
  \centering  
  \vspace{-1mm}
  \includegraphics[width=1.0\linewidth]{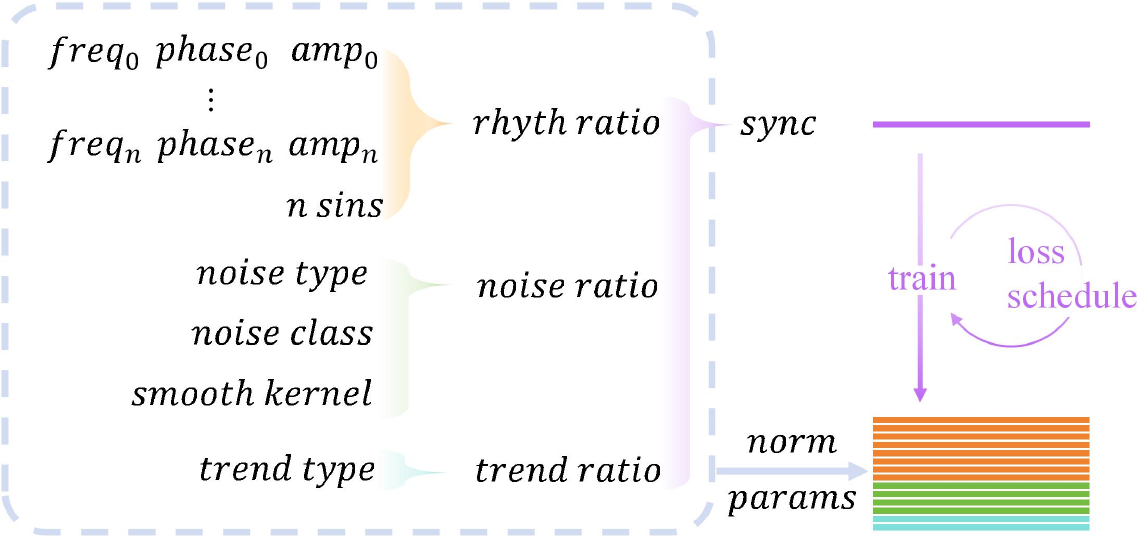}  
  \vspace{-5mm}
  \caption{Normalization of parameters (norm params) used in the data synthesis process. The normalized parameters are organized into a multi-channel matrix, aligning with the sampling window length of the synthetic data.}  
  \label{norm_params}   
  \vspace{-2mm}
\end{figure}

To ensure that a wide range of values and diverse parameter types numbering over a dozen (the total count being dependent on the preset range of the number of sine waves, typically empirically set between 3 and 10), can be effectively fitted as labels for deep models and to mitigate potential interference from different types of loss functions, we have custom-tailored a standardization scheme for all parameters. Ultimately, all values are constrained within the range of -1 to 1 (with some parameters set between 0 and 1). Through broadcasting or interpolation, we map all parameters to a length equivalent to the sampling window of the synthetic data. This process ensures that MRD, TN \& NR, and TI, along with the generated parameters, are organized into a multi-channel matrix of the total parameter count multiplied by the sampling window length, serving as labels. The synthetic data is then utilized as input to train the representation extractor. The parameters contained within the multi-channel normalized parameters are shown in \autoref{norm_params}.

\subsubsection{Representation Extraction Training}

After obtaining the multi-channel normalized parameters (norm params) matrix, which encapsulates the majority of information constituting the synthetic data without containing the data itself, we can train a representation extractor solely on the task of learning from the synthetic data input and the norm params as labels. The norm params encompass various dimensions such as MRD, TN \& NR, and TI, are derived from random sampling and contain diverse continuous random functions. Consequently, the combination of parameters used to generate synthetic data is not limited in quantity, allowing for an infinite variety of corresponding synthetic data. Therefore, training a deep model on the synthetic data to learn the task of MRD, TN \& NR, and TI from norm params makes it nearly impossible for the model to overfit. To achieve optimal extraction of MRD, TN \& NR, and TI within a reasonable timeframe, we adopt a concise linear loss schedule to train the representation extractor. The training process is also depicted in \autoref{norm_params}.

During the training process, in our quest to find the most suitable model architecture for the extraction of MRD, TN \& NR, and TI tasks, we experimented with various model architectures including Bi-LSTM\cite{Abduljabbar2021_bilstm}, DLinear\cite{zeng2023_dlinear}, PatchTST\cite{Nie2023_patchTST}, among others. Based on the visualization results, we selected DLinear as the InfoBoost's Representation Extractor due to its superior visual performance.

\section{Experiments}

To test the generalizability of the synthetic data within the InfoBoost framework and evaluate the representation extraction performance of the Representation Extractor on real-world data, we gathered 35 publicly available time-series datasets. These datasets encompass a wide range of data types, including electroencephalography (EEG) data, epidemiological data, electricity data, cryptocurrency data, traffic data, and meteorological data.

\subsection{Reconstruction of Real Data Solely Trained on Synthetic Data} 
\label{exp_recon}

Because the reconstruction task requires the model to learn from input data and recreate the original data, and evaluate the model’s ability to learn data representations and generalize to new, unseen data, we choose the reconstruction task to validate the generalizability of the synthetic data.

\begin{figure*}[!htbp] 
  \centering
  \includegraphics[width=1.0\linewidth]{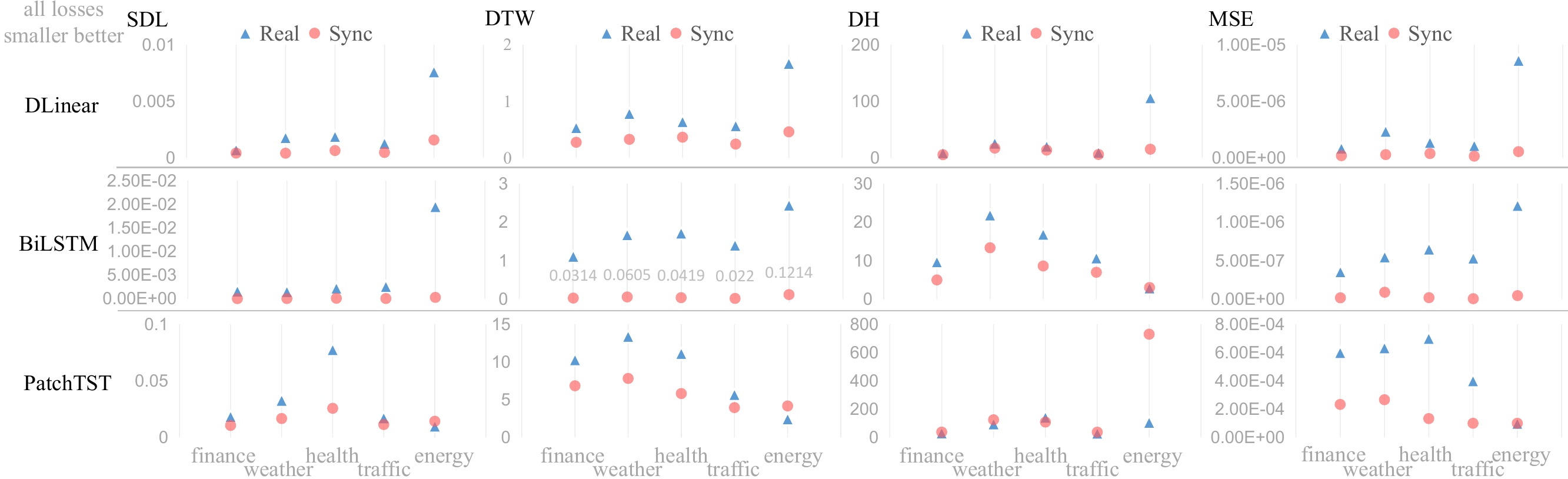}
  \vspace{-5mm}
  \caption{The experimental results in \autoref{sameNumRecon} show that smaller values for all metrics indicate better performance. Here, SDL represents structural dissimilarity loss, DTW represents dynamic time warping distance, and DH represents distance between histograms, MSE represents Mean Squared Error. To demonstrate the reconstruction performance directly across various domains of time-series data, we categorized all tested time-series data into five major groups. We computed the average losses based on each dataset and then calculated the average losses for all datasets within each major group, this allows us to observe relatively reasonable model performance across vastly different types of datasets, even when there is a significant difference in data volume.
}
  \label{ReconExp1}
  \vspace{-2mm}
\end{figure*}

\subsubsection{A Fair Fight with Real Data}
\label{sameNumRecon}

We initially trained a set of models solely on InfoBoost's non-deep learning synthetic data to complete the reconstruction task. And another set of models was trained based on randomly selected portions of real-world datasets to serve as a baseline for completing the reconstruction task on real data. Finally, the remaining real-world datasets were used as a test set to evaluate the performance of the reconstruction models trained on InfoBoost's synthetic data against those trained on real data. To ensure fairness, approximately 70\% of the real-world datasets, specifically 24 datasets, were randomly selected as the baseline training data. Due to the entirely random selection, it is difficult to avoid overlaps between some of the baseline training datasets and the test set in terms of data types, further increasing the difficulty of surpassing the baseline. The selected baseline training set ultimately comprised a specific number (usually around 200,000, depending on random selection) of instances of real time-series data. Similarly, our synthetic data generation also yielded a similar number of instances of synthetic time-series data, which were used to train the model for reconstruction based on InfoBoost’s synthetic data.

We conducted an experiment to evaluate reconstruction performance using four distinct losses. Firstly, the structural dissimilarity loss, derived by subtracting the structural similarity index (SSIM)\cite{venkataramanan2021hitchhikers_SSIM} from 1, is to assess structural variance between the reconstructed and original data. Secondly, the dynamic time warping (DTW)\cite{salvador2007toward_dtw} distance, a metric suitable for measuring similarity between two time series, accomodates time shifts and stretches. Thirdly, the distance between histograms\cite{cha2002measuring_DH} is to gauge dissimilarities between the histograms of the reconstructed and original time-series data, providing insights into their similarity. Lastly, the mean squared error (MSE) loss is a common metric used to quantify differences between the reconstructed and original data. By incorporating these diverse loss functions, we comprehensively evaluated reconstruction performance from various perspectives. To mitigate the impact of model architecture selection on experimental results, we opted for three highly representative diverse model structures: recurrent architecture BiLSTM \cite{Abduljabbar2021_bilstm}, linear networks DLinear \cite{zeng2023_dlinear}, and transformer architecture PatchTST \cite{Nie2023_patchTST}. These were chosen to assess the reconstruction performance on a real data test set after training solely on synthetic data and real data, respectively. The results are shown in \autoref{ReconExp1}.

The results indicate that, except for two scenarios (such as the DH \& DTW value of PatchTST on the Energy class dataset), the performance of the reconstruction model trained on InfoBoost’s synthetic data surpasses that of the reconstruction model trained on real data in the real data test set in 55 out of 60 testing scenarios. It demonstrates the generalization capability of InfoBoost's synthetic data across various types of scenarios. Even in other testing scenarios of the Energy class dataset, although slightly inferior to the performance after training on real data, the performance of the model trained on synthetic data is mostly very close. The reasons for the slightly inferior performance of synthetic data in the PatchTST scenario on the Energy class dataset will be discussed in \autoref{nolimitNumRecon}.

\begin{figure*}[!htbp] 
  \centering
  \includegraphics[width=1.0\linewidth]{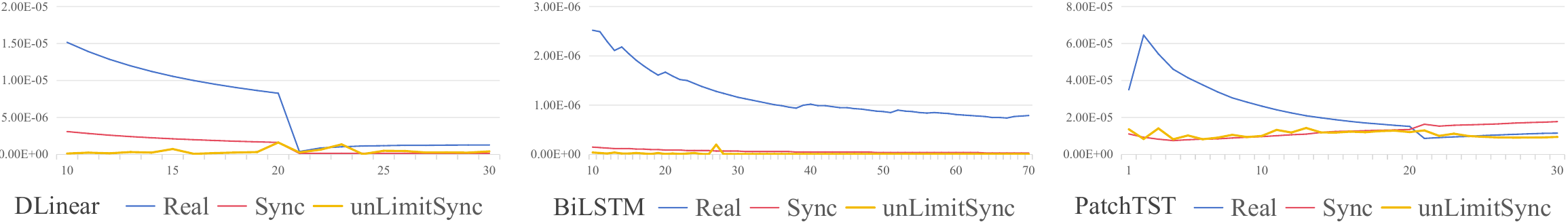}
  \vspace{-7mm}
  \caption{The comparative results, illustrated in the figure, depict the change in the validation set (composed of randomly selected real data) loss for each epoch. Unlike 'Sync', 'unlimitSync' represents the generation of an entirely new synthetic dataset for training at each epoch.
}
  \label{ReconExp2_1}
  \vspace{-2mm}
\end{figure*}

\begin{figure*}[!htbp] 
  \centering
  \includegraphics[width=1.0\linewidth]{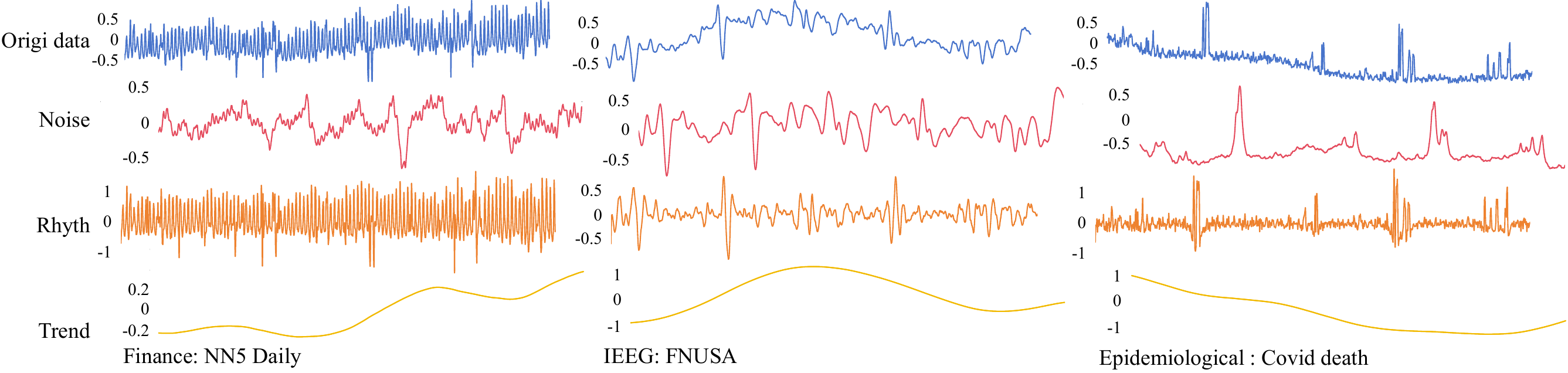}
  \vspace{-7mm}
  \caption{The general feature extractor extracts MRD, TN, NR, and TI information from three distinct and characteristic time-series datasets, following training solely on synthetic data. The ratio information has already been incorporated in a weighted manner across the data scales of each extracted feature after decomposition.
}
  \label{ReconExp3_1}
  \vspace{-2mm}
\end{figure*}

\begin{figure}
  \centering  
  \includegraphics[width=1.0\linewidth]{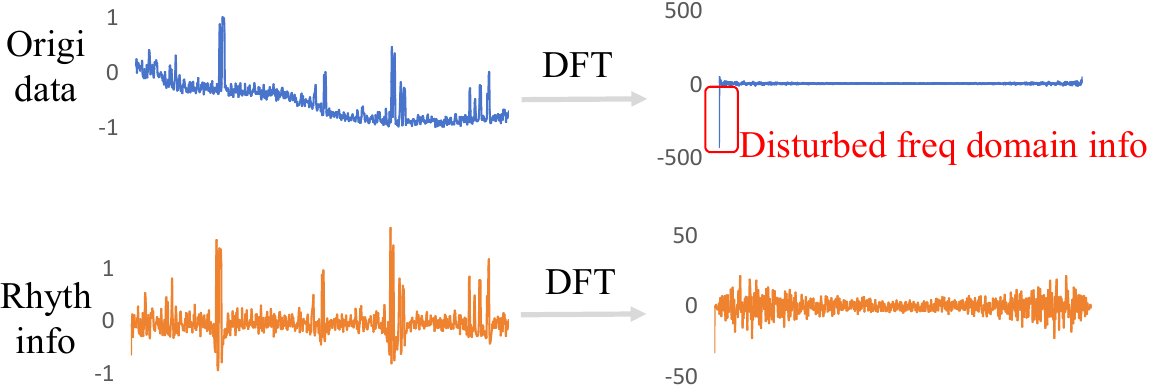}   
  \vspace{-5mm}
  \caption{DFT results in deep learning tasks are based on the input depicted in \autoref{ReconExp3_1}, the COVID death data. Results exhibit an disruption in data's DFT result. After extracting only the rhythmic information, the DFT results for the rhythmic data exhibit higher quality in the frequency domain.}  
  \label{ReconExp3_2} 
  \vspace{-2mm}
\end{figure}

\subsubsection{Giving It All with Synthetic Data}
\label{nolimitNumRecon}

While the reconstruction model based on synthetic data outperforms the model trained on real data, we observe a decline in performance with excessive epochs at a fixed learning rate, particularly in transformer-based models PatchTST. This suggests that although synthetic data enhances generalizability, limiting the size of the training set for the sake of fairness in the experiment may cause the model to overfit to specific features within this subset.

To validate this hypothesis, we modified the reconstruction task in Experiment \autoref{sameNumRecon} to remove the limit on synthetic data quantity. For each completed epoch, a new set of synthetic data was generated to serve as the training data for the reconstruction task. We then compared the change in validation set loss for each epoch between the limited synthetic data reconstruction task and the unlimited synthetic data reconstruction task, using the same model architectures. Based on the experimental results shown in \autoref{ReconExp2_1} and \autoref{validation-losses}, models trained with fixed learning rates and synthetic data replacement at every epoch demonstrates the ability to rapidly and fit well to the validation set with very few epochs, outperforming both the model trained on real data and the model trained on a limited set of synthetic data. Furthermore, as the number of epochs increases, there is no significant increase in validation loss. Instead, the fluctuation in validation loss corresponds to the variation in the generated training data.

\begin{table}[t]  
\label{validation-losses}  
\vskip 0.1in  
\begin{center}  
\begin{small}  
\begin{sc}  
\vspace{-2mm}
DLinear
\begin{tabular}{lcccr}  
\toprule  
Method & minValiLoss & minValiLoss epoch \\  
\midrule  
Real        & $4.0 \times 10^{-7}$ & 21 \\  
Sync        & $1.5 \times 10^{-7}$ & 21 \\  
unlimitSync & $8 \times 10^{-8}$ & 8 \\  
\bottomrule  
\end{tabular}  
BiLSTM
\begin{tabular}{lcccr}  
\toprule  
Method & minValiLoss & minValiLoss epoch \\  
\midrule  
Real        & $7.4 \times 10^{-7}$ & 68 \\  
Sync        & $3 \times 10^{-8}$ & 65 \\  
unlimitSync & $1 \times 10^{-8}$ & 30 \\  
\bottomrule  
\end{tabular}  
PatchTST
\begin{tabular}{lcccr}  
\toprule  
Method & minValiLoss & minValiLoss epoch \\  
\midrule  
Real        & $8.70 \times 10^{-6}$ & 21 \\  
Sync        & $7.52 \times 10^{-6}$ & 5 \\  
unlimitSync & $7.26 \times 10^{-6}$ & 4 \\  
\bottomrule  
\end{tabular}  
\end{sc}  
\end{small}  
\end{center}  
\vskip -0.1in  
\caption{The validation loss for the three test models are presented, each of which is trained on different data sets and minimizes on different epochs. The data in the table indicates that by replacing the training data with new synthetic data at every epoch, the models can achieve better performance in fewer epochs.}
\vspace{-3mm}
\end{table}

\subsection{Case Study of Explicit Eepresentation Extraction}
\label{exp_case}
This section will showcase, through visualizations, the performance of a universal representation extractor trained on synthetic data containing MRD, TN \& NR, and TI information. The visualizations will demonstrate the extraction efficacy of MRD, TN \& NR, and TI across various data types, which is shown in \autoref{ReconExp3_1}. Due to the influence of noise and trend information, the frequency domain information extracted by the commonly employed DFT method in deep learning often experiences a reduction in quality. Therefore, in \autoref{ReconExp3_2}, we present the DFT frequency domain extraction results of rhythmic information based on the InfoBoost feature extractor. This demonstrates the disrupted frequency domain info in origi data's DFT results can be removed of rhythmic information extraction on the frequency domain.


\section{Conclusion}

In this study, we have introduced a unique approach, marking the first to simultaneously fulfill the requirements of a universal time-series data synthesis method that does not rely on real data or deep learning, and a universal time-series data representation decomposition and extraction method that does not require fine-tuning on real data. Most notably, our method empowers models trained in the absence of real data information to outperform those trained on real data across almost all tested datasets. This achievement opens up a new path for future time-series data analysis and modeling, as well as a new direction for unsupervised or self-supervised learning.

\end{document}